\documentclass[sigconf, dvipsnames,table,xcdraw,preprint]{acmart}

\usepackage{epsf}
\usepackage{framed}
\usepackage{url}
\usepackage{graphicx}
\usepackage{dblfloatfix}
\usepackage{enumitem}
\usepackage{hyperref}
\usepackage{tabularx}
\usepackage{xr-hyper}
\usepackage[ruled,vlined,linesnumbered]{algorithm2e}
\usepackage{booktabs, makecell, tabularx}
\usepackage{amsmath}
\usepackage{mathtools}
\usepackage[labelsep=period]{caption}
\usepackage{soul} 
\usepackage{graphicx}
\usepackage{caption}
\usepackage{subcaption}
\usepackage[figuresright]{rotating}
\usepackage{float}
\usepackage{multirow}
\usepackage{tikz}
\usepackage{lscape}
\usepackage{tabu}
\usepackage{tabularx}
\usetikzlibrary{shapes, arrows.meta, positioning}
\SetKwComment{Comment}{/* }{ */}

\makeatletter
\def\@copyrightspace{\relax}
\makeatother

\setcopyright{none}
\renewcommand\footnotetextcopyrightpermission[1]{} 
\begin{document}

\title{CBAG: An Efficient Genetic Algorithm \\ for the Graph Burning Problem}


\author{Mahdi Nazeri}
\affiliation{%
  \department{Department of Electrical and Computer Engineering,}
  \institution{Isfahan University of Technology}
  \city{Isfahan 84156-83111} 
  \country{Iran} 
}
\email{m.nazeri@ec.iut.ac.ir}

\author{Ali Mollahosseini}
\affiliation{%
  \department{Department of Electrical and Computer Engineering,}
  \institution{Isfahan University of Technology}
  \city{Isfahan 84156-83111} 
  \country{Iran} 
}
\email{mollahosseini@ec.iut.ac.ir}

\author{Iman Izadi}
\affiliation{%
  \department{Department of Electrical and Computer Engineering,}
  \institution{Isfahan University of Technology}
  \city{Isfahan 84156-83111} 
  \country{Iran} 
}
\email{iman.izadi@iut.ac.ir}



\newcommand{\algoone}[1][]{
\begin{algorithm}[!t]
\caption{Calculating the $\nu$-th row of the middle matrix}

\label{algo:middle}
\DontPrintSemicolon
\SetKwInput{Input}{Input~}
\SetKwInput{Output}{Output~}
\Input{~Vertex $\nu$, APSP matrix $Dist$ of size $N \times N$, BFS traversal sequence $S_{\nu}$, and parents of vertices in array $Parent$ of size $N$}
\Output{~Middle vertices between $\nu$ and others stored in the $\nu$-th row of $MiddleMatrix$}
\Begin{
	$i \leftarrow 0$\;
	$j \leftarrow 0$\;
	\While{$i \leq N$}{
	    $u \leftarrow S_{\nu}[i]$\;
	    $mid \leftarrow S_{\nu}[j]$\;
    	\uIf{$Dist[\nu, u] \% 2 = 1$}{
    	    $parentMid \leftarrow MiddleMatrix[\nu, Parent[u]]$\;
    	    $MiddleMatrix[\nu, u] \leftarrow parentMid$\;
    	    $i \leftarrow i + 1$\;
    	}
    	\uElseIf{$Dist[\nu, mid] = Dist[mid, u]$ {\bf and} $Dist[\nu, mid] + Dist[mid, u] = Dist[\nu, u]$
    	}{
    	    $MiddleMatrix[\nu, u] \leftarrow mid$\;
        	$i \leftarrow i + 1$\;
    	}
    	\Else{
    	    $j\leftarrow j + 1$\;
    	}
	}
}
\end{algorithm}
}

\begin{abstract}
Information spread is an intriguing topic to study in network science, which investigates how information, influence, or contagion propagate through networks.
Graph burning is a simplified deterministic model for how information spreads within networks. The complicated NP-complete nature of the problem makes it computationally difficult to solve using exact algorithms. Accordingly, a number of heuristics and approximation algorithms have been proposed in the literature for the graph burning problem.
In this paper, we propose an efficient genetic algorithm called Centrality BAsed Genetic-algorithm (CBAG) for solving the graph burning problem. Considering the unique characteristics of the graph burning problem, we introduce novel genetic operators, chromosome representation, and evaluation method. In the proposed algorithm, the well-known betweenness centrality is used as the backbone of our chromosome initialization procedure.
The proposed algorithm is implemented and compared with previous heuristics and approximation algorithms on
15 benchmark graphs of different sizes. Based on the results, it can be seen that
the proposed algorithm achieves better performance in comparison to the previous state-of-the-art heuristics.
The complete source code is available online and can be used to find optimal or near-optimal solutions for the graph burning problem. 
\end{abstract}



\maketitle
\pagestyle{plain}

\section{Introduction}
Models that describe the process of spreading information or influence through networks are of high interest and have been studied in a number of domains \cite{kempe2003maximizing}.
For instance, an application of these models can be seen in political campaigns.
During such campaigns, candidates deliver speeches in order to broadcast their thoughts to the public. Suppose a candidate delivers one speech per day for $b$ consecutive days.
As a result of each day's speech, the candidate reaches out to exactly one new group of people.
This new group is considered to be \emph{informed}.
Meanwhile, previously informed groups share the candidate's thoughts with their neighboring groups, and they will be deemed informed as well.
Two groups are considered neighbors if they interact consistently.
As delivering these speeches are quite costly and time-consuming, candidates seek to inform all target groups with a minimum number of speeches. Therefore, it is necessary for candidates to deliver their speeches in an optimal manner (i.e., an optimal sequence of groups).

In \cite{kempe2003maximizing}, another application has been explained. Consider a satellite needs to spread a piece of information to all nodes in a network. The satellite itself can transmit information to a single node at each time instant. In addition, when a node receives information, it informs all of its neighbors at the next time instant. Due to performance considerations, it is essential to inform all nodes within a minimum number of time instants. Accordingly, the satellite should transmit the information to an optimal sequence of nodes.

As a model for the spread of social influence or information, graph burning has been introduced by Bonato et al. \cite{bonato2014burning} in 2014.
Simply stated, in the graph burning model, each node in a network spreads information to its immediate neighbors after it receives that information (i.e., it burns its neighbors after it is burned). This flow of information continues until all the nodes receive the information (i.e., they are all burned).
The graph burning problem, as introduced in \cite{bonato2014burning}, is to find a sequence of nodes to give the information, in order to inform the entire network in the least amount of time.
In~\cite{bonato2014burning}, the burning number parameter was presented as a measure of spread speed and offered an upper bound conjecture on the burning number that recently proved asymptotically in \cite{norin2022burning}.

Parameterized complexity of graph burning was studied in \cite{kare2019parameterized, kobayashi2022parameterized}. The theoretical aspects of the problem such as algorithms, bounds, and complexity for different classes of special graphs have been greatly investigated.
For detailed information refer to the recently published survey on graph burning \cite{bonato2021survey}.
Approximation algorithms were studied in \cite{bessy2017burning, kamali2020burning, bonato2019approximation, bonato2019bounds}.
Approximation algorithms with approximation ratio of 3 were proposed in \cite{bessy2017burning, bonato2019approximation}. A 2-approximation algorithm was given for trees in \cite{bonato2019approximation}.
In addition to approximation algorithms, heuristics have been considerably studied and proposed in \cite{simon2019burn,farokh2020new,gautam2022faster,vsimon2019heuristics}.

Šimon et al. proposed three heuristics in \cite{vsimon2019heuristics}: Maximum Eigenvector Centrality Heuristic (MECH), Cutting Corners Heuristic (CCH), and Greedy Algorithm with Forward-Looking Search Strategy Heuristic (GFSSH). The first heuristic, MECH, selects the vertex with the highest eigenvector centrality value and appends it to the burning sequence. Starting with an empty burning sequence, this procedure is repeated until a valid burning sequence is generated.
The second heuristic, CCH, employs an algorithm to select a set of vertices that are considered to be corners. These corners are used along with eigenvector centrality in order to determine candidate (central) vertices. The candidate vertices are then passed to the weighted aggregated sum product assessment (WASPAS) algorithm in order to select the next fire source. The third heuristic, GFSSH, considers less promising vertices in addition to the WASPAS algorithm output. At each step, a fire source is chosen among these vertices using forward-looking search algorithms such as A*. This contributes for a more complete search space and usually obtains better results.

In their subsequent work on graph burning, Šimon et al. \cite{simon2019burn} implemented MECH with 30 different centrality measures other than eigenvector centrality. In their study, the centrality measures were implemented and compared on several different datasets. Their results suggest that barycenter centrality and closeness centrality are the most effective centrality measures for generic networks, whereas betweenness centrality is the most effective measure when applied to geometric networks (e.g. mobile networks).

Recently, Gataum et al. \cite{gautam2022faster} proposed three heuristics based on eigenvector centrality. These three heuristics are Backbone Based Greedy Heuristic (BBGH), Improved Cutting Corners Heuristic (ICCH), and Component Based Recursive Heuristic (CBRH).
They have implemented and compared their algorithms with other heuristics on SNAP and Network Data Repository datasets \cite{leskovec2014snap, rossi2015network}. Reported results show that their algorithms achieve the same results but are faster than other algorithms such as three heuristics proposed by Šimon et al. in \cite{vsimon2019heuristics}.

In the literature, graph burning has been associated with a couple of well-known problems. Vertex k-center \cite{garcia2018local} is a related problem, in which there is no order in the burning process, and all sources fire simultaneously at start. In addition, as a complementary but distinct version of the graph burning problem, the Firefighter problem \cite{finbow2009firefighter} is also worth mentioning. At each step of the Firefighter problem the firefighter protects one node from fire in order to reduce the spread of fire in the network.

Even moderately sized networks contain thousands of nodes and tens of thousands of edges. Given the fact that graph burning is NP-complete \cite{bessy2017burning}, it is computationally challenging and typically impossible to find a global optimum solution in a reasonable amount of time for most real-world applications. As a result, heuristic and approximation algorithms are used to find near-optimal solutions. The Genetic Algorithm (GA) is a population-based stochastic optimization technique which is inspired from natural selection mechanisms \cite{sivanandam2008genetic}. Genetic algorithms have been successfully applied to a variety of optimization problems across a broad range of fields, including engineering, logistics, management, and economics \cite{alam2020genetic}.

In this paper, Centrality BAsed Genetic-algorithm (CBAG), an efficient genetic algorithm approach is proposed to solve the graph burning problem.
Initially, the precalculation step is performed in order to calculate and store some information that is necessary for the subsequent steps.
The proposed genetic algorithm uses centrality measures to generate the initial generation of chromosomes.
Chromosomes are then evolved using specially designed crossover and mutation operators. 
To evaluate the proposed algorithm, CBAG was tested on 15 benchmark graphs of different sizes. The results show that, in general, CBAG outperforms the previous heuristics and approximation algorithms.

The rest of this paper is organized as follows. In the next section, the formulation of the graph burning problem is presented. Section~\ref{algorithm} provides a detailed explanation of the genetic algorithm proposed in this research. Section~\ref{disconnected} discusses how CBAG procedure is applied to disconnected graphs. Section~\ref{results} presents our results and compares them with previous studies. Section~\ref{conclusion} discusses our conclusions and potential future directions.

\section{Problem Formulation} \label{formulation}

Graph burning is a simplified graph-based model for deterministic information or influence spread.
It has been introduced by Bonato et al. in \cite{bonato2014burning} and is defined as follows.
Consider a finite, simple, and undirected graph $G$ with vertex set $V(G)$ with cardinality $N$ and edge set $E(G)$ with cardinality $M$. The graph burning process consists of $b$ discrete steps from $t_{0}$ to $t_{b - 1}$.
At each step, each vertex is either burned or unburned (i.e. informed or uninformed). 
All vertices are initially unburned at step $t_0$.
At each step, one new unburned vertex is set on fire and burned by an exogenous agent.
We call these vertices \emph{fire sources}.
In addition, the vertices that were burned in the previous step, spread the fire to their neighbors and burn them as well.
Once a vertex is burned, it remains burned throughout the process.
This process continues until all vertices are burned.
The objective is to find an optimal sequence of vertices serving as fire sources in order to burn all vertices within a minimum number of steps, which is the same as the minimum number of fire sources.

A sequence of fire sources is considered to be a \emph{burning sequence} in the case that after firing the last fire source (i.e., after the last step of burning process) no vertex remains unburned. Burning sequences with a minimal length are considered to be optimal. The length of an optimal burning sequence of graph $G$ is called \emph{burning number} and is denoted by $BN(G)$.

For any two vertices $u$ and $v$ in $V(G)$, $d(u, v)$ is the number of edges in a shortest path between $u$ and $v$ in graph $G$. Given positive number $k$, the $k$-th closed neighborhood of $v$ is defined as the set $\{u: d(u, v) \le k\}$ and is denoted by $N_k[v]$. As shown in \cite{bonato2014burning,bonato2016burn}, for a given graph $G$, vertex sequence $[v_1, v_2, \dots, v_k]$ forms a valid burning sequence if and only if for every pair of $i$ and $j$ that \linebreak $1\le i < j \le k$; $d(v_i, v_j) \ge j - i$ is met, and the following condition holds:
$$N_{k-1}[v_1] \cup N_{k-2}[v_2] \cup \dots \cup N_{0}[v_k] = V(G)$$

\begin{figure}[!t]
    \centering
    \begin{tikzpicture}
      [scale=1,auto=center,every node/.style={circle,fill=brown!5, draw=black!90,scale=1,minimum size=0.65cm}]
      \node[line width=0.2mm] (a2)  at  (1,1)     {2};  
      \node[line width=0.2mm] (a3)  at  (2,1)     {3};
      \node[line width=0.2mm] (a4)  at  (3,1)     {4};
      \node[line width=0.2mm] (a5)  at  (4,1)     {5};      
      \node[line width=0.2mm] (a6)  at  (5,1)     {6};
      \node[line width=0.2mm] (a7)  at  (6,1)     {7};
      \node[line width=0.2mm] (a1)  at  (4,2)     {1};
      \node[line width=0.2mm] (a8)  at  (4,0)     {8};
      \node[line width=0.2mm] (a10) at  (4,-1)   {10};
      \node[line width=0.2mm] (a12) at  (4,-2)   {12};
      \node[line width=0.2mm] (a9)  at  (3,-1)    {9};
      \node[line width=0.2mm] (a11) at  (5,-1)   {11};
      \draw[very thick, black!90] (a2) -- (a3);
      \draw[very thick, black!90] (a3) -- (a4);
      \draw[very thick, black!90] (a4) -- (a5);
      \draw[very thick, black!90] (a5) -- (a6);
      \draw[very thick, black!90] (a6) -- (a7);
      \draw[very thick, black!90] (a1) -- (a5);
      \draw[very thick, black!90] (a1) -- (a4);
      \draw[very thick, black!90] (a1) -- (a6);
      \draw[very thick, black!90] (a5) -- (a8);
      \draw[very thick, black!90] (a6) -- (a8);
      \draw[very thick, black!90] (a10) -- (a8);
      \draw[very thick, black!90] (a10) -- (a9);
      \draw[very thick, black!90] (a10) -- (a11);
      \draw[very thick, black!90] (a10) -- (a12);
    \end{tikzpicture}
    \caption{Vertex sequences [4, 10, 7] and [5, 10, 2] both are optimal burning sequences. [2, 10, 1, 7] is also a valid burning sequence, but it is not optimal. [3, 8, 12] is not a valid burning sequence since vertices 7, 9, and 11 remain unburned.}
    \label{fig:example network}
\end{figure}
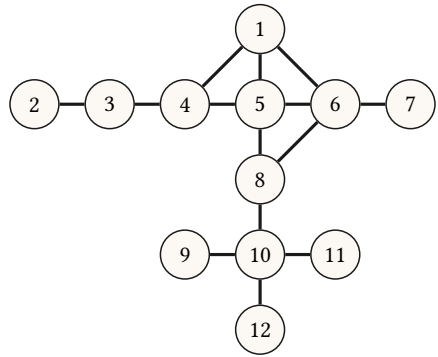

As an example, consider the graph $G$ presented in Figure~\ref{fig:example network}. The sequence of vertices [4, 10, 7] is a valid burning sequence for the graph, since all vertices are burned at the end of the last step. This burning sequence is an optimal burning sequence, as there is no valid burning sequence of length less than three. This is due to the fact that $G$ has a maximum degree (i.e. maximum number of neighbors of any vertex) of four, which means that the first source of fire can burn no more than four vertices. Furthermore, the second fire source cannot burn any vertices aside from itself. Consequently, no burning sequence of length two is able to burn more than (1 + 4) + (1 + 0) = 6 vertices. In addition, the optimal burning sequence is not necessarily unique, as the sequence [5, 10, 2] is also a valid burning sequence of length three.

\section{The Proposed Genetic Algorithm} \label{algorithm}

In this paper, the Centrality BAsed Genetic-algorithm (CBAG) is introduced in order to solve the decision version of the graph burning problem. Given a burning length $b$ in addition to the graph $G$, the objective of the decision version is to determine if any valid burning sequence of length at most $b$ exists. The proposed algorithm $\textsc{CBAG}(G, b)$ either finds such a solution and returns \emph{true}, or fails to find one and returns \emph{false}.
If it returns \emph{true}, then a valid burning sequence of length at most $b$ is found and can be obtained from the algorithm. The algorithm returns \emph{false} if it does not find any solution within a reasonable number of generations. It should be noted that failure of the algorithm doesn't necessarily imply that a burning sequence of length at most $b$ does not exists. A reasonable approach for finding the burning number of graph $G$ (i.e. length of an optimal burning sequence of graph $G$) is to use binary search on $b$ in order to find the minimum burning length which the algorithm returns \emph{true}.

Based on our observations, the binary search over \textsc{CBAG}$(G, b)$ does not significantly increase the computational complexity of the algorithm. This is due to the fact that while the precalculation part of the algorithm is computationally intensive, it does not need to be computed more than once as it depends only on the graph $G$ and not on any other parameters such as $b$. Also, valid burning sequences are usually generated more quickly for $b$ values greater than $BN(G)$. 

\begin{figure*}[ht]
    \centering 
    \includegraphics[page=1]{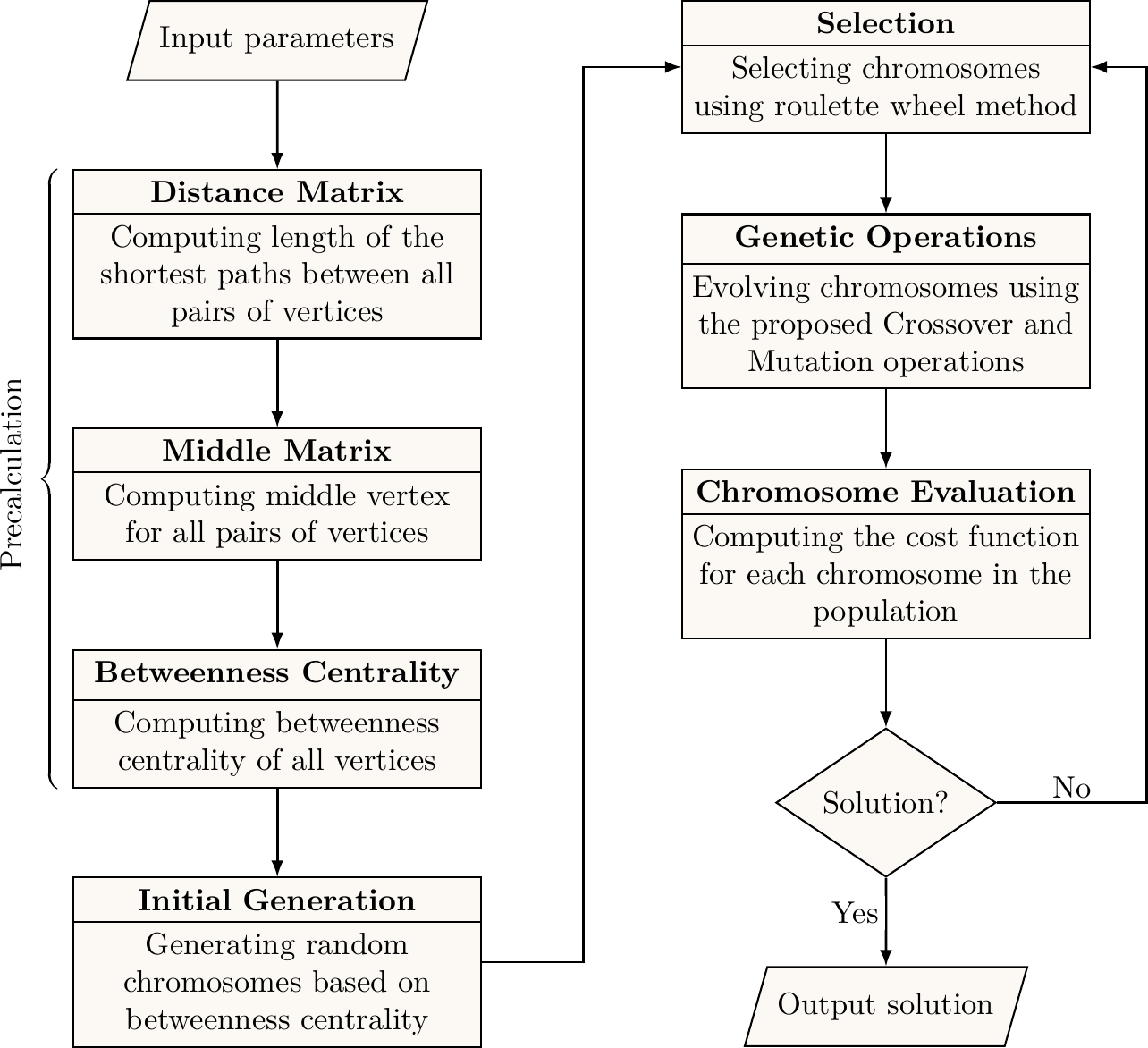}
    \caption{Flowchart of the proposed genetic algorithm.}
    \label{fig:flowchart}
\end{figure*}

As an effective optimization technique, we applied genetic algorithm to the graph burning problem. Flowchart of the proposed algorithm for finding optimal or near-optimal solutions is shown in Figure~\ref{fig:flowchart}. Initially, the algorithm precalculates some information that it needs in the next steps, then initializes a population of randomly generated chromosomes. This population of chromosomes then goes through a self-development process. Throughout this process, the fitness of each chromosome is evaluated. Subsequently, fitter chromosomes are selected to evolve using genetic operations such as crossover and mutation. This procedure continues until a valid burning sequence of length at most $b$ is found.
 
\subsection{Precalculation}

There are a few elements that should be precalculated before the proposed genetic algorithm can proceed. Several parts of the algorithm depend on knowing the length of the shortest path between each pair of vertices. These values are computed using an All-Pairs Shortest Path (APSP) algorithm. Additionally, a middle vertex for each pair of vertices has to be computed. Vertex $m$ is called the middle vertex of two vertices $i$ and $j$, if the following conditions hold: vertex $m$ lies on a shortest path between $i$ and $j$, and $|d(i, m) - d(m, j)|\le1$.

The proposed algorithm begins by computing the APSP matrix. For each vertex of the graph, its distance from all other vertices is computed using BFS traversal algorithm. This procedure can be computed in $\Theta(N\times(N+M))$. Although many algorithms have been proposed for the APSP problem \cite{madkour2017survey}, the current approach is used since it is simple to implement and performs better on non-dense graphs, which is the case for most real-world networks \cite{melancon2006just}. 

Once the APSP matrix is computed, the algorithm determines the middle vertex of each pair of vertices. For each vertex $\nu$, a BFS traversal starting at vertex $\nu$ is performed, and the order in which the vertices are visited is preserved in the sequence $S_{\nu}$.
The parent of each vertex in that BFS tree is also maintained.
Algorithm~\ref{algo:middle} computes the middle vertices between $\nu$ and others using the two pointer technique. After running this algorithm for every vertex $\nu$ in the graph, the middle vertex for each pair of vertices is computed. Since the precalculation procedure runs BFS traversal $N + N$ times and Algorithm~\ref{algo:middle} is linear in time, the total complexity of the precalculation is $\Theta(N\times(N+M))$.

The betweenness centrality of each vertex of the graph is also computed in this step of the algorithm. A fast algorithm proposed by Brandes \cite{brandes2001faster} can be used to compute the betweenness centrality of all vertices in $\Theta(NM)$. However, several approximation algorithms have been developed for betweenness centrality in order to achieve more efficient computations. A benchmark for betweenness centrality approximation
algorithms is provided in \cite{matta2019comparing}. In this work, we used the approximation algorithm proposed in \cite{brandes2007centrality}.

\algoone{}

\subsection{Chromosome representation}
It is important to understand the different characteristics of fire sources in different positions of the burning sequence. The fire sources at the beginning of the burning sequence are fired earlier, and thus have a high potential for spreading the fire widely across the graph. For instance, when the length of the burning sequence is greater than the graph's radius, an optimal choice for the first fire source (i.e. any of Jordan centers) would burn all vertices by itself.
In contrast, the last fire source is only able to burn itself, and the next to the last fire source is only able to burn itself and its neighbors. In an optimal burning sequence, we expect the early fire sources to:
\begin{itemize}
  \item Spread the fire widely across the graph.
  \item Have high centrality with respect to many centrality measures.
  \item Depend on global features of the graph instead of prior choices for fire sources.
  \item Be distant from each other in order to spread the fire through different parts of the graph.
\end{itemize}

In the proposed algorithm, each chromosome is a sequence of fire sources and is represented by an ordered list of vertices.
All chromosomes have a length of \emph{ChrSize}, which is determined before the algorithm begins and remains unchanged throughout the execution.
Different chromosomes represent different burning sequences of length $b$, where $b \geq \emph{ChrSize}$. However, chromosomes do not contain all $b$ fire sources of the burning sequences they represent. Each chromosome only contains the prefix with length \emph{ChrSize} of the burning sequence it represents. Considering that each chromosome is an ordered list of vertices with length \emph{ChrSize}, the vertex in position $i$ ($1 \leq i \leq \emph{ChrSize}$) is the fire source that is fired at time instant $t_{i-1}$. The remaining burning sequence for each chromosome contains $b  - \emph{ChrSize}$ fire sources. A search algorithm is used to determine the optimal choice for the remaining fire sources of each chromosome independently. This is discussed further in \ref{evaluation}~Evaluation.

\begin{figure*}[!t]
    \centering
    \includegraphics[page=1]{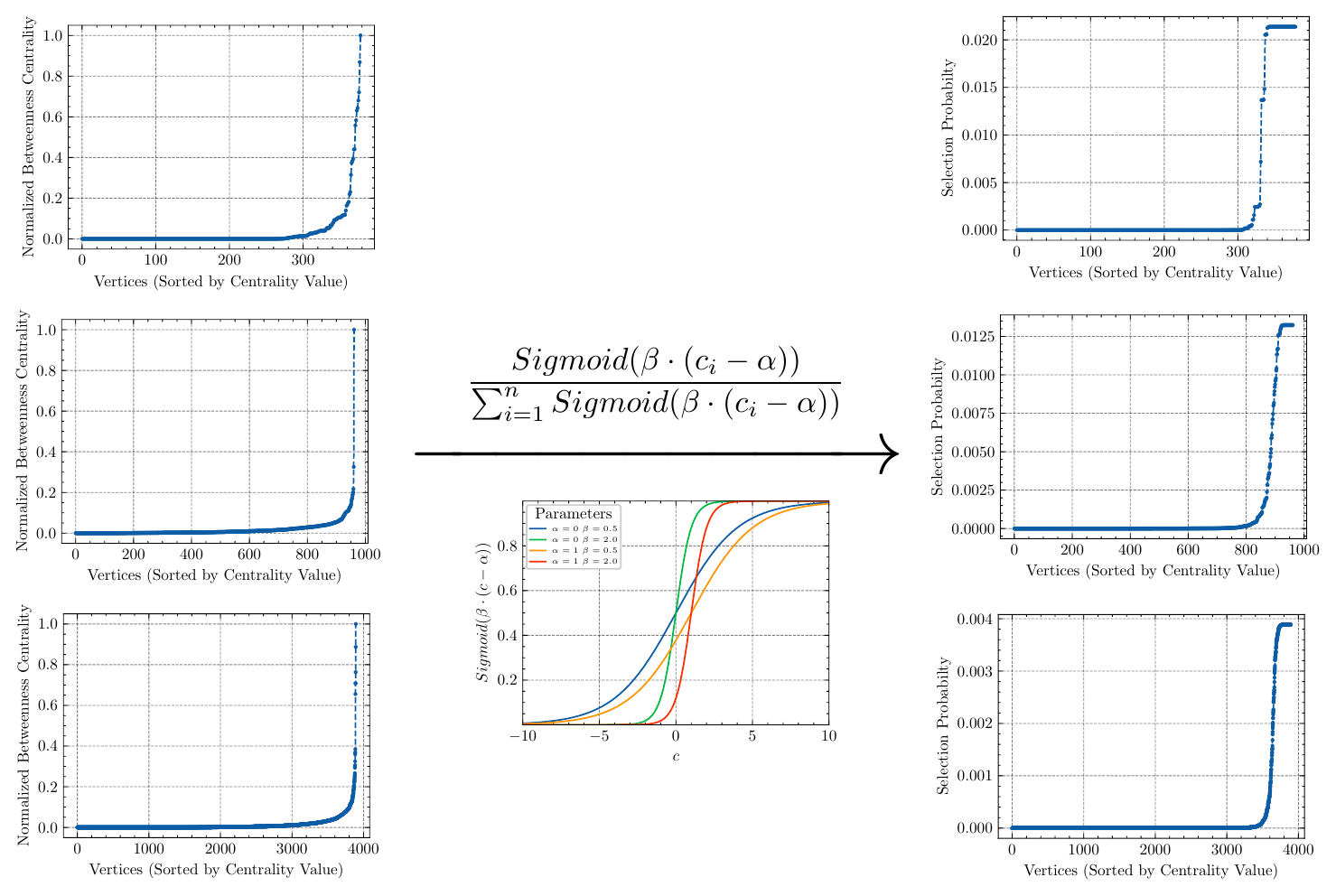}
    \caption{On the left is the sorted betweenness centrality of three different graphs: Netscience, Reed98, and TVshow from top to bottom, respectively. On the right is the selection probability of vertices in the introduced sampling process.
    Same hyperparameters $\alpha = 0.05$ and $\beta = 200$ are used for each of the three graphs.}
    \label{fig:Selection}
\end{figure*}

The advantages of this approach are twofold. 
First, the genetic algorithm only has to determine a prefix of an optimal burning sequence, while the rest is determined using search techniques. As a result, the genetic algorithm has a significantly reduced search space.
Second, optimal choices for the late fire sources are expected to heavily depend on prior choices for the earlier fire sources. This is in contrast to early fire sources, which are more dependent on the graph's global features. Therefore, excluding late fire sources from chromosomes leads to more independent fire sources in each chromosome. This is helpful for genetic operations to alter different parts of each chromosome independently and effectively.

\subsection{Chromosome initialization} \label{Chromosome initialization}
The first generation of population is created by randomly generated chromosomes. 
A random procedure is introduced to generate new chromosomes. 
The procedure generates one new chromosome per each execution and therefore needs to be run for \emph{PopSize} number of times in order to initialize the first generation.
Every execution of this procedure contains \emph{ChrSize} steps. At each step, one vertex is randomly selected as the next fire source and is placed in the leftmost available position of the chromosome. Specifically, in each step $i$ from $0$ to $\emph{ChrSize} - 1$, the procedure selects one vertex to set on fire at time instant $t_i$ of the burning process. This vertex is placed in position $i + 1$ of the chromosome.

The vertices are selected using a non-uniform sampling process. The probability of selecting each vertex depends on its centrality. Formally, given normalized centrality values $c_i$ ($1 \le i \le N$) and hyperparameters $\alpha, \beta$, the probability of selecting vertex $i$ is:
$$
\frac{Sigmoid(\beta\cdot(c_i - \alpha))}{T}
$$
where $T$ is the sum of all numerators and is calculated as follows:
$$
T = \sum_{i = 1}^{N}{Sigmoid(\beta\cdot (c_i - \alpha))}
$$
An illustration of this procedure is provided in Figure~\ref{fig:Selection} for several benchmark graphs.

In order to spread fire in different parts of graph, we expect early fire sources in any optimal chromosome to be distant from each other. At each step of the above procedure, vertices with distance less than \emph{MinDist} to any of the previously selected fire sources in prior steps are excluded from the sampling process. This exclusion can be simply made by setting the corresponding centrality values to zero. At the end of the procedure, these centralities are set back to their initial values and would not affect the generation of subsequent chromosomes.

The hyperparameter \emph{MinDist} should be chosen appropriately. Choosing small values for \emph{MinDist} results in close proximity of sampled fire sources, and therefore may leave some parts of the graph far from fire and reduces the chromosome's fitness. On the other hand, choosing large values can cause the algorithm to get stuck, as at some step there may be no vertex that is sufficiently distant from the previously sampled fire sources. The following parameter tuning mechanism is used in order to prevent the procedure from getting stuck: whenever the procedure gets stuck, meaning there is no vertex that is sufficiently distant from previously sampled fire sources, the \emph{MinDist} parameter is reduced by one. This reduction may be repeated several times. At the end of the process, the \emph{MinDist} parameter sets back to its initial value and would not affect the generation of subsequent chromosomes.

In this work, betweenness centrality is used as the centrality measure.
The performance of different centrality measures for the graph burning problem is thoroughly studied in \cite{simon2019burn}. 
In their analysis, eigenvectors and betweenness centrality were the most effective measures of centrality. Eigenvector and betweenness centrality are implemented and compared as the centrality measure which is used in our algorithm.
Using betweenness centrality, the proposed algorithm achieved significantly better performance on previous benchmarks in the literature.

\begin{figure}[!t]
    \centering
    \begin{tikzpicture}
    \node[] (BCCenter)
    {\includegraphics[scale=0.075]{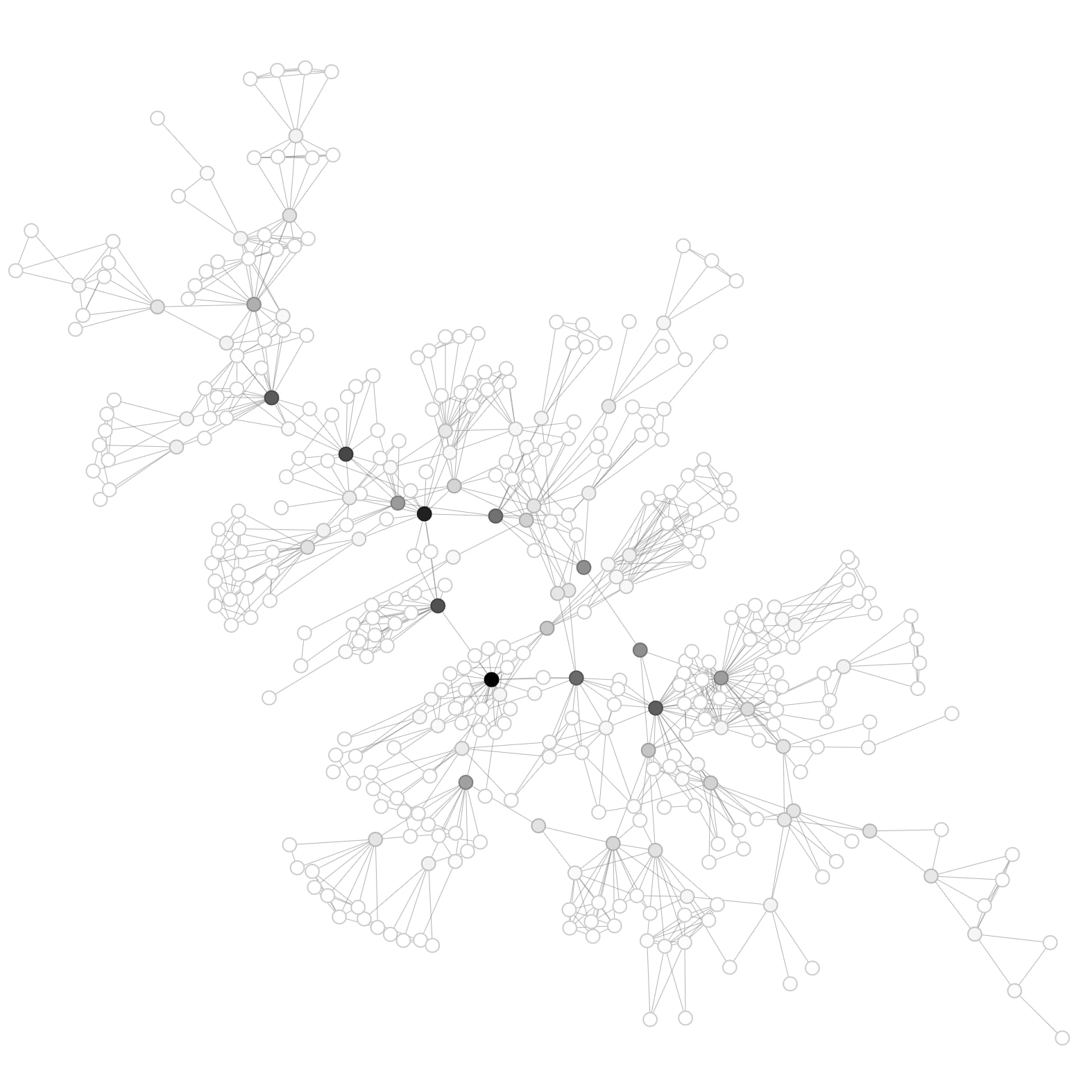}};
    \node[right= -0.2cm of BCCenter] (EVCenter)
    {\includegraphics[scale=0.075]{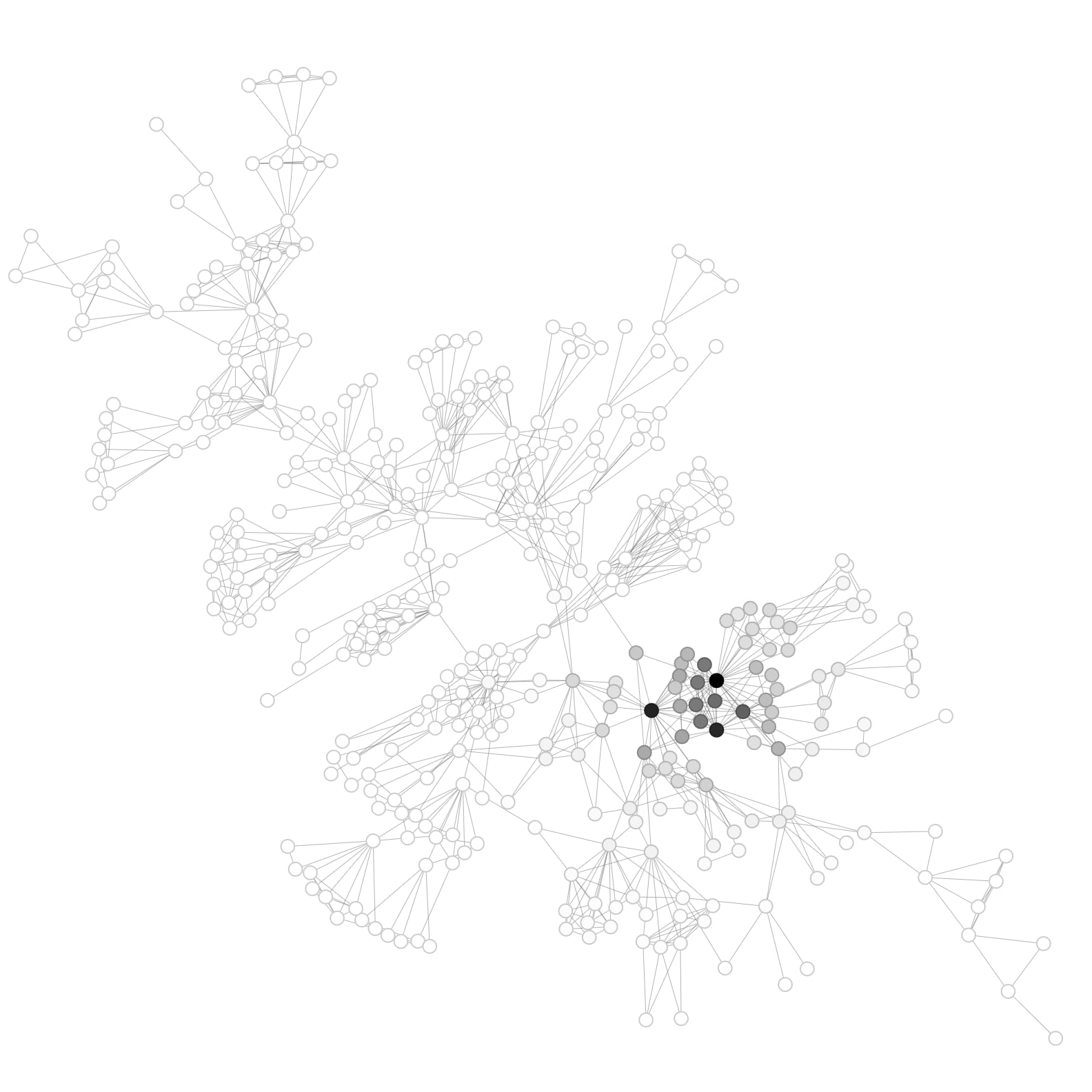}};
    \end{tikzpicture}
    \caption{Betweenness (left) and eigenvector (right) centrality of the Netscience network. Darker vertices indicate higher centrality values. In contrast to the eigenvector centrality, vertices with high betweenness centrality values are distributed across the entire network.}
    \label{fig:centrality}
    \vspace{-0.3cm}
\end{figure}

According to our observations on the benchmark graphs, betweenness centrality has the following advantages:
First, betweenness centrality makes a clear distinction between central vertices and the others. In the benchmark graphs, a few percent of vertices have high centrality, whereas more than 80\% of them have zero or near-zero centrality values. Second, vertices with high betweenness centrality are widely scattered across the entire graph. This allows us to generate chromosomes consisting of central vertices that are distant from each other and are distributed in different parts of the graph. Whereas in most centrality measures, vertices with high centrality are usually clustered in a few parts of the graph.
Refer to Figure~\ref{fig:centrality} for an illustration.

\subsection{Evaluation} \label{evaluation}
The fitness function used in our algorithm to evaluate chromosomes is described in this subsection. As defined earlier, a vertex is considered to be burned if it is set on fire at any time instant of the burning process. At the end of the burning process, the distance from vertex $i$ to the nearest burned vertex is called \emph{burning distance} of vertex $i$ and is denoted by $d_i$. Consequently, the burning distance of any burned vertex is zero. For any valid burning sequence, all vertices are burned during the burning process and thus have a burning distance of zero.

The cost of a burning sequence is defined as the sum of squared burning distances of all vertices. Formally, the cost of a burning sequence $S=(v_1, v_2, \dots, v_{b})$ of length $b$ is formulated as
$
\sum_{i=1}^{N}{{d}_i^2}
$
where $d_i$ can be calculated as:
\begin{equation*}
d_i= 
\begin{cases}
  \:\: 0 & \text{if vertex $i$ is burned,} \\
  \min\limits_{1 \leq j \leq b}{dist(i, v_j) - (b - j)} & \text{otherwise.}
\end{cases}
\end{equation*}
The minimum distance of each pair of vertices is precalculated in the \emph{DistMatrix} and can be retrieved in $O(1)$.

As discussed earlier, each chromosome only contains the first \emph{ChrSize} fire sources of the burning sequence it represents. Therefore, to evaluate each chromosome, it is necessary to determine the optimal choice for the remaining fire sources of the chromosome, and then calculate the cost function of the obtained burning sequence. The cost of the chromosome is calculated after the remaining fire sources are completed by each ordered subset of unburned vertices with length $b - \emph{ChrSize}$. The minimum cost among all these burning sequences is considered as the cost of the chromosome and is denoted by $Cost(C)$ where $C$ is the given chromosome.

A vertex is called unburned with respect to some chromosome when it is not burned by first \emph{ChrSize} fire sources in a burning process with $b$ steps. In other words, a vertex is called unburned with respect to chromosome $C=(v_1, v_2, \dots, v_{ChrSize})$ if:

$$
\min\limits_{1 \leq j \leq ChrSize}{dist(i, v_j) - (b - j)} > 0
$$
In order to obtain a valid burning sequence, the remaining fire sources of the chromosome must be chosen such that every unburned vertex is burned by one of these fire sources.

The number of possible ways to complete the burning sequence of an individual chromosome using a subset of unburned vertices \emph{unburned} is in $\Theta(|unburned|^{b - ChrSize})$. This number can be quite large and lead to computationally heavy evaluation, specifically for chromosomes with ineffective choices of early fire sources. These chromosomes are not expected to lead to valid burning sequences due to their non-optimal early fire sources. Therefore, for faster computations, chromosomes with number of unburned vertices greater than \emph{skipNumber} are skipped from the evaluation process and assumed to have a large cost of \emph{INFCost}. In this work, we used a complete brute-force search over unburned vertices in order to determine the remaining fire sources of each chromosome. However, different approaches such as using incomplete search heuristics are more computationally efficient and can be utilized instead.

\subsection{Selection}
Selection is an important part of genetic algorithms that determines which chromosomes participate in genetic operations such as crossover and mutation. In the proposed algorithm, the roulette wheel selection method is used as our selection operation. In roulette wheel selection, each chromosome participates in genetic operations with a probability proportional to its fitness. Thus, chromosomes with higher fitness are more likely to be selected for genetic operations. An efficient roulette wheel selection method with $O(1)$ time complexity is presented in \cite{LIPOWSKI20122193}.

It is important to note that fitter solutions in our population are chromosomes with lower cost values. However, they should be selected with higher probability. In this regard, the selection probability of chromosome $C_i$ is calculated as:

$$
P_i = \frac{W_i}{\sum_{j=1}^{PopSize}{W_j}}
$$
where $W_i$ is the selection weight of chromosome $C_i$ and is equal to $\frac{1}{Cost(C_i) + 1}$.
\subsection{Genetic operations}
Genetic algorithms search the solution space using crossover and mutation operations. Crossover and mutation are key elements for exploring and exploiting the search space. Although many different crossover and mutation operators have been proposed in the literature, they are not effectively applicable in this work due to the unique characteristics of the graph burning problem and the proposed genetic algorithm. Therefore, we propose new genetic operators for the graph burning problem based on the following key properties:
\begin{itemize}
    \item The order of fire sources in the chromosomes are not altered by these operations. This is due to the fact that fire sources in different positions of chromosome have different characteristics.
    \item The vertices in the neighborhood of each fire source are explored by these operations.
    \item The amount of exploration of each vertex is directly related to the centrality value of that vertex.
\end{itemize}

\begin{figure}[!t]
    \centering
    \includegraphics[scale=0.79]{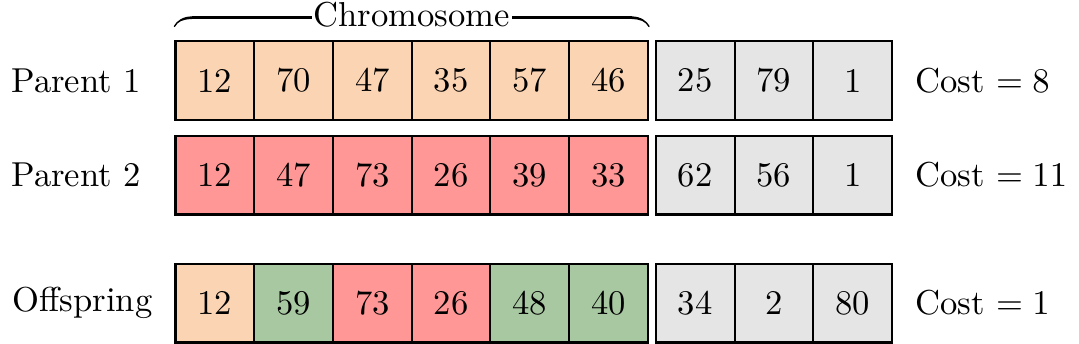}
    \caption{An example of the proposed crossover operation for path graph $P_{81}$ and $ChrSize = 6$ is visualized. Each of the burning sequences is divided into two parts: the fire sources in the chromosome which are shown in colors, and an optimal choice for the remaining fire sources determined by the evaluation function in gray. The fire sources of the first and second chromosomes are highlighted in orange and red, respectively. The green fire sources indicate the middle vertex between the parent fire sources.}
    \label{fig:crossover}
    \vspace{-0.3cm}
\end{figure}

\subsubsection{Crossover}
Crossover is an operation in which the chromosomes of two or more parents are combined into a new chromosome called offspring. In this work, we present a novel crossover operator that produces an offspring by merging the fire sources of two parents. The $i$-th $(1 \leq i \leq ChrSize)$ fire source of the offspring chromosome is chosen with equal probability among the following three options: 
\begin{enumerate}[label=\arabic*.]
    \item The $i$-th fire source of the first parent.
    \item The $i$-th fire source of the second parent.
    \item The middle vertex between fire sources at the $i$-th position of the parent chromosomes. (These fire sources must be in the same connected component in order to have any middle vertices.)
\end{enumerate}
Using this procedure, the properties of both parents are partially retained, and the search space is further explored. An example of the proposed crossover operator is shown in Figure~\ref{fig:crossover}.

\begin{figure}[!t]
    \centering
    \includegraphics[scale=0.68]{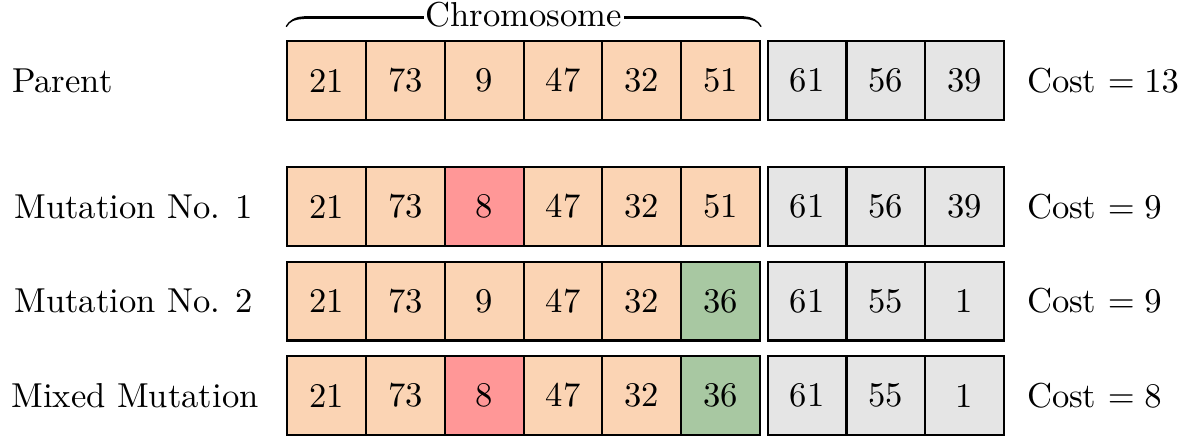}
    \caption{The figure illustrates an example of the proposed mutation operations for path graph $P_{81}$ and $ChrSize = 6$. Mutation No. 1 uses neighboring vertices to mutate the parent chromosome. Mutation No. 2 replaces each fire source in \emph{MutationSet} with some random vertex in the same component. The red and green fire sources are mutated using the first and second mutation operators, respectively. Gray fire sources are determined using the evaluation function and would not participate in mutation.}
    \label{fig:mutation}
    \vspace{-0.3cm}
\end{figure}

\subsubsection{Mutation}
Mutation operators are used to explore the search space and avoid trapping in local optima. Two different mutation operators are introduced and employed in the proposed algorithm. Each of the mutation operators initially selects a set of fire sources of the input chromosome and stores them in \emph{MutationSet}. Each position of the input chromosome is added to \emph{MutationSet} with the equal probability \emph{MutationProb}.

The first mutation replaces each fire source $v \in \emph{MutationSet}$ with one of its neighbors $u \in N(v)$ with the equal probability $\frac{1}{|N(v)|}$. The general idea of this mutation is to explore nearby vertices that do not necessarily have high centrality values.

The second mutation replaces each fire source in \emph{MutationSet} with a random vertex in the same connected component of the graph. This random vertex is selected using a non-uniform sampling process in which the probability of selecting each vertex depends on its centrality. This is identical to the sampling process introduced in Subsection \ref{Chromosome initialization}. This mutation helps to explore the problem space and avoid being trapped in local optima. 
An example of the proposed mutation operators is illustrated in Figure~\ref{fig:mutation}.

\section{Burning of disconnected graphs} \label{disconnected}
The graph burning problem is not restricted to connected graphs and can be generalized to disconnected graphs easily.
In this case, at least one fire sources must be placed in each connected component of the graph in order to burn the graph completely.
The proposed algorithm performs well on disconnected graphs due to its following properties: 

\begin{itemize}
    \item Betweenness centrality is normalized for each connected component independently. In addition, for components with less than three vertex, the centrality of each vertex is considered to be one. This normalization helps for initial chromosomes to contain fire sources from more different components.
    \item The proposed crossover and mutation operators are also applicable to disconnected graphs and perform well.
    \item As distance of each pair of vertices in different components is considered to be infinite, reasonable values for \emph{MinDist} parameter lead most of initial chromosomes to contain fire sources from all components.
\end{itemize}

\section{Results} \label{results}
We have implemented the proposed genetic algorithm in C++ utilizing the Boost library. The complete source code of CBAG is available at the GitHub repository\footnote{\href{https://github.com/aloomya/CBAG}{https://github.com/aloomya/CBAG}}.
The most comprehensive benchmark in the literature is reported by Gautam et al. in \cite{gautam2022faster}. They implemented and compared their heuristics with approximation algorithms \cite{bonato2019approximation}, and heuristics proposed by Farokh et al. \cite{farokh2020new} and Šimon et al \cite{vsimon2019heuristics}. Their benchmark contains a number of graphs from The Network Data Repository \cite{rossi2015network}, Stanford large network dataset collection (SNAP Datasets) \cite{leskovec2014snap}, and randomly generated graphs using Barabasi-Albert and Erdos-Renyi models. In order to show the effectiveness of the proposed algorithm, we compare the results of CBAG with previously proposed heuristics and approximation algorithms on their benchmark graphs.

Table~\ref{tab:parameters} presents the genetic parameters and their values. The same hyperparameters are used for all benchmark instances. Table~\ref{tab:machine spec} provides the technical specifications of the machine that is used for the analysis. For our results to be comparable, a machine with similar performance to the machine used in their study is utilized. 
Table \ref{tab:performance comparison} compares the results of CBAG with previous heuristics and approximation algorithms. The comparison of their execution time is provided in Table \ref{tab:time comparison}. Each reported result represents an average of 10 executions of the proposed genetic algorithm.

Based on the results, it can be seen that CBAG achieves better solutions for several benchmark graphs in comparison to the previously proposed heuristics. For other graphs in the benchmark, the algorithm achieves the same results as the best previously proposed heuristic. For most of the benchmark graphs, we expect the result of our algorithm to be the burning number and therefore may not be improved.

\section{Conclusion} \label{conclusion}
In this work, we proposed an efficient genetic-based algorithm for the graph burning problem. Considering the unique characteristics of the graph burning problem, three essential components have been introduced throughout the development of the algorithm: the initialization procedure, the evaluation function, and novel crossover and mutation operators. In addition, an explanation of how the algorithm operates on disconnected graphs is provided.

The genetic algorithm proposed in this paper (CBAG) has proven to be both effective and efficient.
We tested CBAG on 15 benchmark graphs and compared our results with previous state-of-the-art heuristics.
In each benchmark instance, CBAG achieved better or equal results. Additionally, the execution time of CBAG was comparable to the fastest proposed heuristic. 
In conclusion, CBAG can be used to find optimal or near-optimal solutions for the graph burning problem.

\begin{table}[h]
    \centering
    \caption{Parameters of the genetic algorithm used for the analysis.}
    \begin{tabularx}{\columnwidth}
    {>{\hsize=1\hsize\linewidth=\hsize}X
    >{\hsize=1\hsize\linewidth=\hsize}X}
        \hline
        Parameter & Value\\
        \hline
        Chromosome Size & $b - 3$\\
        Minimum Distance & Equal to Chromosome Size\\
        Skip Number & 20\\
        Population Size & 300\\
        Maximum Generations & 500\\
        Crossover Population & 500\\
        Mutation Probability & 0.1\\
        Alpha & 0.05\\
        Beta & 200\\
        \hline
    \end{tabularx}
    \label{tab:parameters}
\end{table}

\begin{table}[h]
    \centering
    \caption{Specification of the machine used for the analysis.}
    \begin{tabularx}{\columnwidth}{>{\hsize=1.25\hsize\linewidth=\hsize}X
>{\hsize=0.75\hsize\linewidth=\hsize}X}
        \hline
        Parameter & Value\\
        \hline
        CPU Model Name & Intel Xeon\\
        CPU Family & Haswell\\
        CPU Frequency & 2.20 GHz\\
        CPU Cores & 2\\
        RAM & 16 GB\\
        OS & Ubuntu 18.04.3\\
        \hline
    \end{tabularx}
    \label{tab:machine spec}
\end{table}

\begin{landscape}
    \begin{table}[!ht]
         \centering
        \caption{Comparison of the obtained burning lengths of the proposed approximations in \cite{bonato2019approximation}, heuristics proposed in \cite{vsimon2019heuristics, gautam2022faster} and CBAG.}
        \begin{tabularx}{\columnwidth}{l l l l X X X X X X X l}
            \hline
            Network source & Name & |V| & |E| & 3APRX~\cite{bonato2019approximation} & 2APRX~\cite{bonato2019approximation} & GFSSH~\cite{vsimon2019heuristics} & CCH~\cite{vsimon2019heuristics} & BBGH~\cite{gautam2022faster} & ICCH~\cite{gautam2022faster} & CBRH~\cite{gautam2022faster} & CBAG \\
            \hline
            Network data & Netscience &  379 & 914 & 12 & 10 & 7 & 7 & 7 & 7 & 7 &  \textcolor{red}{\textbf{6}} \\
            repository & Polblogs & 643 & 2K & 9 & 10 & 6 & 6 & 6 & 6 & 6 & \textcolor{red}{\textbf{5}}\\
            & Reed98 & 962 & 18K & 6 & 8 & 4 & 4 & 4 & 4 & 4 & 4\\
            & Mahindas & 1258 & 7513 & 9 & 8 & 5 & 5 & 5 & 5 & 5 & 5\\
            & Cite-DBLP & 12.6K & 49.7K & 120 & 82 & 41 & 41 & 41 & 41 & 41 & 41\\
            SNAP Dataset & Chameleon & 2.2K & 31.4K & 9 & 10 & 6 & 6 & 6 & 6 & 6 & 6\\
            & TVshow & 3.8K & 17.2K & 18 & 16 & 10 & 10 & 10 & 10 & 10 & \textcolor{red}{\textbf{9}}\\
            & Ego-Facebook & 4K & 88K & 9 & 6 & 4 & 4 & 4 & 4 & 4 & 4\\
            & Squirrel & 5K & 198K & 9 & 10 & 6 & 6 & 6 & 6 & 6 & 6\\
            & Politician & 5.9K & 41.7K & 12 & 12 & 7 & 7 & 7 & 7 & 7 & 7\\
            & Government & 7K & 89.4K & 9 & 10 & 6 & 6 & 6 & 6 & 6 & 6\\
            & Crocodile & 11K & 170K & 12 & 10 & 6 & 6 & 6 & 6 & 6 & 6\\
            & Gemsec-Deezer (HR) & 54K & 498K & 12 & 12 & 7 & 7 & 7 & 7 & 7 & 7\\
            Randomly & Barabasi-Albert & 1K & 3K & 6 & 8 & 4.9 & 4.9 & 4.9 & 4.9 & 4.9 & \textcolor{red}{\textbf{4.2}}\\
            Generated & Erdos-Renyi & 1K & 6K & 6 & 8 & 5 & 5 & 5 & 5 & 5 & 5\\
            \hline
        \end{tabularx}
        \label{tab:performance comparison}
    \end{table}
    
    \begin{table}[!b]
        \raggedleft 
        \caption{Comparison of execution time of the heuristics proposed in \cite{vsimon2019heuristics, gautam2022faster} and CBAG.}
        \begin{tabularx}{\columnwidth}{l l X X X X X X X X}
            \hline
            Network Source & Name & |V| & |E| & GFSSH~\cite{vsimon2019heuristics} & CCH~\cite{vsimon2019heuristics} & BBGH~\cite{vsimon2019heuristics} & ICCH~\cite{gautam2022faster} & CBRH~\cite{gautam2022faster} & 
            CBAG\\
            \hline
            Network Data & Netscience & 379 & 914 & 2m & 3s & <1s & <1s & 1s & <1s\\
            Repository & Polblogs & 643 & 2K & 3s & 3s & <1s & 1s & 2s & <1s\\
            & Reed98 & 962 & 18K & 5s & 6s & 3s & 3s & 5s & <1s\\
            & Mahindas & 1258 & 7513 & 6s & 7s & <1s & 3s & 23s & <1s\\
            & Cite-DBLP & 12.6K & 49.7K & 3m 8s & 3m 20s & 39s & 22s & 2m & 22s\\
            SNAP Dataset & Chameleon & 2.2K & 31.4K & 25s & 27s & 20s & 16s & 16s & 2s\\
            & TVshow & 3.8K & 17.2K & 30s & 26s & 7s & 22s & 15s & 18s\\
            & Ego-Facebook & 4K & 88K & 1m & 1m & 17s & 16s & 22s & 5s\\
            & Squirrel & 5K & 198K & 3m 5s & 2m & 40s & 34s & 1m 40s & 14s\\
            & Politician & 5.9K & 41.7K & 1m & 52s & 14s & 32s & 17s & 7s\\
            & Government & 7K & 89.4K & 1m 13s & 1m 17s & 20s & 50s & 32s & 14s\\
            & Crocodile & 11K & 170K & 5m & 3m 17s & 2m 36s & 42s & 4m & 40s\\
            & Gemsec-Deezer (HR) & 54K & 498K & 1h 20m & 49m & 2m 36s & 7m & 47m & 17m 30s\\
            Randomly & Barabasi-Albert & 1K & 3K & 1m 10s & 1m & 10s & 10s & 20s & <1s\\
            Generated & Erdos-Renyi & 1K & 6K & 1m 40s & 1m 30s & 10s & 5s & 30s & <1s\\
            \hline
        \end{tabularx}
        \label{tab:time comparison}
    \end{table}
\end{landscape}

\clearpage
\onecolumn
\bibliographystyle{ACM-Reference-Format}
\bibliography{bibliography}
\end{document}